\documentclass{article}

     \PassOptionsToPackage{numbers}{natbib}

    \usepackage[preprint]{neurips_2023}

\usepackage[utf8]{inputenc} 
\usepackage[T1]{fontenc}    
\usepackage{url}            
\usepackage{booktabs}       
\usepackage{amsfonts}       
\usepackage{nicefrac}       
\usepackage{microtype}      
\usepackage{xcolor}         

\usepackage[pdftex]{graphicx}
\usepackage{booktabs} 
\usepackage{balance}
\usepackage{amsmath, amssymb}
\usepackage[subrefformat=parens]{subcaption}
\usepackage{appendix}

\newtheorem{dfn}{Definition}
\newcommand\figref[1]{Figure~\ref{fig:#1}}

\makeatletter
\newcommand\figcaption[1]{\def\@captype{figure}\caption{#1}}
\newcommand\tblcaption[1]{\def\@captype{table}\caption{#1}}
\makeatother

\title{Is Self-Supervised Pretraining Good for Extrapolation in Molecular Property Prediction?}

\author{%
  Shun Takashige\\
  The University of Tokyo\\
  \texttt{shige@eidos.ic.i.u-tokyo.ac.jp} \\
  \And
  Masatoshi Hanai \\
  The University of Tokyo\\
  \texttt{hanai@ds.itc.u-tokyo.ac.jp} \\
  \AND
  Toyotaro Suzumura \\
  The University of Tokyo\\
  \texttt{suzumura@ds.itc.u-tokyo.ac.jp} \\
  \And
  Limin Wang\\
  The University of Tokyo\\
  \texttt{lwang@eidos.ic.i.u-tokyo.ac.jp} \\
  \And
  Kenjiro Taura \\
  The University of Tokyo\\
  \texttt{tau@eidos.ic.i.u-tokyo.ac.jp} \\
}

\begin{document}

\maketitle

\begin{abstract}
  The prediction of material properties plays a crucial role in the development and discovery of materials in diverse applications, such as batteries, semiconductors, catalysts, and pharmaceuticals.
  Recently, there has been a growing interest in employing data-driven approaches by using machine learning technologies, in combination with conventional theoretical calculations.
  In material science, the prediction of unobserved values, commonly referred to as \textit{extrapolation}, is particularly critical for property prediction as it enables researchers to gain insight into materials beyond the limits of available data.
  However, even with the recent advancements in powerful machine learning models, accurate extrapolation is still widely recognized as a significantly challenging problem.
  On the other hand, \textit{self-supervised pretraining} is a machine learning technique where a model is first trained on unlabeled data using relatively simple pretext tasks before being trained on labeled data for target tasks.
  As self-supervised pretraining can effectively utilize material data without observed property values, it has the potential to improve the model's extrapolation ability.
  In this paper, we clarify how such self-supervised pretraining can enhance extrapolation performance.
  We propose an experimental framework for the demonstration and empirically reveal that while models were unable to accurately extrapolate absolute property values, self-supervised pretraining enables them to learn relative tendencies of unobserved property values and improve extrapolation performance.
\end{abstract}

\section{Introduction \label{sec:intro}}
%

The prediction of material property values is essential for the development and discovery of materials across a broad range of applications such as batteries, semiconductors, catalysts, and pharmaceuticals.
The prediction is typically conducted for a large set of candidate materials, aiming to identify those that meet the desired property requirements.
As the computational cost of calculating material properties based on physical simulations, such as density functional theory (DFT) or coupled cluster (CC), can be excessively high, data-driven approaches that employ surrogate models trained on a subset of simulation results from the candidate materials have attracted much attention. 

The existing work on the prediction of material property values has mainly focused on \textit{interpolation} problems, where the assumption is that the training data and test data are \textit{independent and identically distributed} (i.i.d.).
However, in the material development and discovery, \textit{extrapolation} problems are practically much more important, i.e., training data and test data are assumed to have different distributions.
One of the ultimate goals in material development is to discover a material with a completely unseen physical property, which would be achieved only through extrapolation.

\begin{figure*}
  \centering
  \begin{minipage}{0.49\hsize}
    \centering
    \includegraphics[width=0.99\hsize]{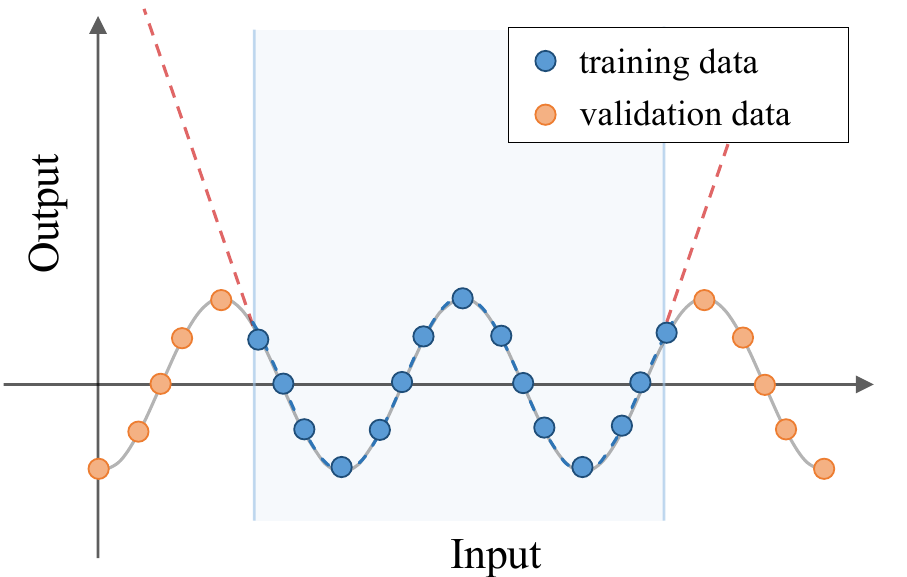}
    \label{fig:ood_input}
  \end{minipage}
  \hfill
  \begin{minipage}{0.49\hsize}
    \centering
    \includegraphics[width=0.99\hsize]{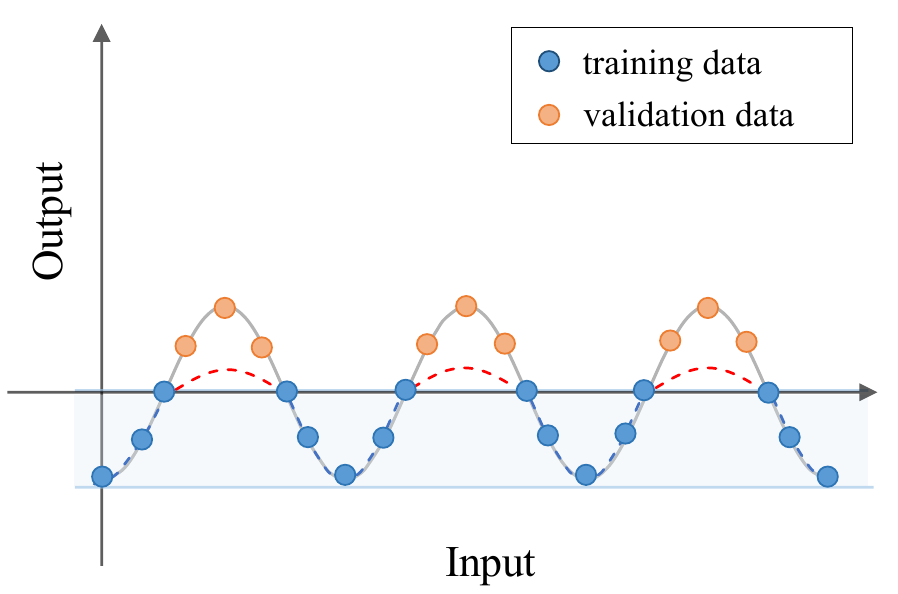}
    \label{fig:ood_output}
  \end{minipage}
  \caption{Two examples of extrapolation and distributions in the case of predicting an underlying function $g(x) = \sin(x)$. The red dotted line is the expected prediction line. \textbf{Left}: based on input features, the training distribution and extrapolation are determined. \textbf{Right}: based on output labels, the training distribution and extrapolation is determined.}
  \label{fig:ood_example}
\end{figure*}

The formation of data distribution is significant in formulating the extrapolation problem, and the existing researches can be categorized into two distinct approaches: those based on \textit{input features} and those on \textit{output labels}.
\figref{ood_example} is one of the simplest cases of extrapolation and shows the difference between them when predicting $g(x) = \sin(x)$.
In the former approach, the data forms distribution based on input features as most researches ~\cite{Hendrycks2019-kl, Peng2018-ab, Li2018-iy, Peng2017-lm, Adila2022-bv, Hendrycks2020-ij, Na2022-rp,Wu2022-ur, Yu2022-ch, Zhang2022-lr,Li2022-ql, Yang2022-iq} work on it.
Thus, the training data and test data include different types of input, such as different colors or sizes of image data.
In the latter approach, the data forms distribution based on (continuous) labels~\cite{Yang2021-mr}.
Thus, the training data and test data include different types of labels, such as different ages in face image data.
The typical classification task always uses the input-feature-based data distribution for the extrapolation since its labels are discrete, making it impossible to predict unknown labels.

In this paper, we focus on the extrapolation problem formulated by the label-based data distribution as material developments have a strong need to predict unobserved physical property values, i.e., unseen labels.
Our research differs from the existing label-based approarch~\cite{Yang2021-mr} in two key aspects.

First, we define the extrapolation differently in a manner that is more appropriate for the application of predicting material property values.
As \cite{Yang2021-mr} has mainly focused on addressing label imbalance issues in image recognition tasks,
the extrapolation region is defined as the special case of the imbalance region where the label data is completely absent.
In contrast, we define an extrapolation region as one that is not included within the maximum and minimum label values.
This definition is more relevant in material property value prediction, where extreme values (e.g., exceptionally large or small values) are crucial.

Second, our study puts emphasizes the utilization of self-supervised pretraining for the extrapolation, whereas \cite{Yang2021-mr} proposes a training method to correct an imbalance of label data.
Our research highlights the unique situation in material property prediction, where numerous unlabeled data points are available with known input structural data, yet the material properties themselves remain unknown.
Therefore, for the utilization of such unlabeled data, we leverage self-supervised pretraining, which enables the model to learn useful features from the input structural data without relying on explicit label information.

In this paper, we first formulate extrapolation for material property prediction considering the distribution of labels.
Next, we analyze the extrapolation ability of existing model based on the formulation.
Finally, we demonstrate the effectiveness of several self-supervised pretraining strategies in improving the extrapolation ability.

To summarize the main contributions, this paper
\begin{itemize}
  \item re-defines an extrapolation problem in material property value prediction. 
  \item reveals that existing models could not predict over a certain value in extrapolation.
  \item reveals that self-supervised pretraining improves the results of extrapolating the relative tendencies to some extent, although property value itself could not be answered exactly.
  \item reveals that the utilization of validation data when pretraining does contribute the performance improvement.
\end{itemize}

\section{Background\label{sec:background}}


\subsection{Extrapolation \label{sec:extrapolation}}

%

Extrapolation generally means predicting data sampled from unseen distributions or domains.
The general definition of extrapolation is described in the following way according to \cite{xu2021neural}.
Firstly, this paper lets $\mathcal{X}$ be the domain of interests such as the structural data of materials.
A target function $g \colon \mathcal{X} \to \mathbb{R}$ generates labels from the inputs.
A label $y \in \mathbb{R}$ and an input $\boldsymbol{x}\in \mathcal{X}$ have a relationship of $y = g(\boldsymbol{x})$.
Given that $\mathcal{D}$ is the support of certain training distribution, training data are expressed as $\{(\boldsymbol{x}_{i}, y_{i})\}^{n}_{i=1} \subset \mathcal{D}$.
Note that as an input $\boldsymbol{x}\in \mathcal{X}$ can be an arbitrary format such as a graph, various definitions can be considered for the data region and distribution.
Extrapolation aims to learn the distribution $\mathcal{X} \setminus \mathcal{D}$ by minimizing the extrapolation error $\mathbb{E}_{\boldsymbol{x} \sim \mathcal{X} \setminus \mathcal{D}}\left[ \ell \left( f(\boldsymbol{x}), g(\boldsymbol{x}) \right) \right]$. Note that $f \colon \mathcal{X} \to \mathbb{R}, \ell \colon \mathbb{R} \times \mathbb{R} \to \mathbb{R}$ is a model and a loss function.
Also, it is recognized as a difficult task as models are poor at learning non-linearity outside the training distributions.
This issue was examined by Xu et al.(2021) \cite{xu2021neural} proving that MLP with ReLU as an activation function could not converge nonlinear function outside training distribution.

\subsection{Molecular Property Prediction}

In molecular property prediction, the input is a molecular graph and the output is a continuous label.
Since a set of graphs can be represented as $\mathcal{G}=(V, E)$ using a set of nodes $V$ representing atoms, and edges $E$ representing bonds, the problem is to learn the relationships $g\colon \mathcal{G} \to \mathbb{R}$.
To model it, GNNs and transformer-based models are extensively developed \cite{Liu2022-sx, masters2022gps++, Ying2021-ye}. 

Although they mainly work in i.i.d. assumption, some research tackled extrapolation, or out-of-distribution generalization in molecular property prediction.
Then, they are specified according to the formation of data distributions: one based on input features or based on out labels.

Most previous studies \cite{Yang2022-iq, Li2022-ql} defined extrapolation by dividing data distributions based on input features.
For example, Yang et al. (2022) \cite{Yang2022-iq} use scaffolds and sizes of graphs to decide distributions.
Although Chan et al. (2021)\cite{Chan2021-wf} formed distributions by output labels, its aim is generating data rather than prediction. 
Whereas the definition based on input features has the advantage of making models robust to a wide variety of inputs, it doesn't meet the needs of finding material extraordinary property value. 
To realize it, a new definition of extrapolation based on output labels in molecular property prediction is expected.



\subsection{Self-Supervised Learning on Graph \label{sec:ssp}}

Self-supervised learning is a method of training a model using only input data without labels.
Since self-supervised learning does not use labels, it has been widely studied for utilizing unlabeled data.
Considering the fact that there are plenty of molecules whose target property values are unknown, it is a powerful tool for our research.
The following three types of pretext tasks are related to self-supervised learning on graphs.

\textbf{Attribute Level Task}.
A popular method for node-level pretext tasks is a node and edge masking problem.
This task is a classification problem, and the model predicts true attributes from the hidden inputs by masking.
These masking tasks were first developed in the field of natural language processing, such as BERT \cite{Devlin2018-dz}, and have been applied to graphs as well.
Through this task, encoders can gain effective information related to graph attributes. \cite{Rong2020-hd, Feng2022-fx}

\textbf{Structure Level Task}.
The task handles structures in graphs and extracts important information regarding the connectivity between nodes without considering the attributes.
Structures strongly influence the characteristics of molecules because electrons involved in the bonds have an important role to determine the property values.
For example, the number of rings in a graph can be a useful indicator for grasping molecules.

\textbf{Attribute + Structure Level Task}.
The task is mainly to predict the properties of subgraphs with attributes and structures.
This often needs to prepare a set of specific graphs in advance for training objectives, e.g., whether the specific graphs are included in a given input graph.
Most of them perform graph matching between subgraphs of inputs and specific graphs in a set of graphs called motif vocabulary \cite{Rong2020-hd}.
The motif vocabularies are often created using domain knowledge in chemistry.


\section{Experimental Framework\label{sec:framework}}
We have to define extrapolation to fit the aim of finding molecules with high or low property values.
For the new definition, distribution based on output labels should be described clearly, as we expect models to predict the property values, or output labels, outside that of training distribution.
It can be realized since the molecular property prediction is a regression task and its labels are continuous.
In addition to this, self-supervised pretraining is considered to be a good solution for improving extrapolation performance.
Since the technique can utilize graph structures whose labels are unknown when training and may improve the problems of non-linearity according to \cite{xu2021neural}, we have to demonstrate its impact on extrapolation.

To evaluate the performance considering the above points, we propose an experimental framework.
Firstly, the formulation of label-based data distribution and extrapolation in material science are introduced.
Then, the hypothesis regarding self-supervised pretraining and the utilization of unlabeled data are described.
Finally, details of experiments for verifying them are explained including the way to pretraining.

\subsection{Problem Definition \label{sec:problem}}

Instead of the traditional definition, we define extrapolation by dividing data distributions based on output labels.
We formulate this definition using the same notations in section \ref{sec:extrapolation}, and it can be described as follows.
\begin{dfn}[Extrapolation in our study]
  We assume the model $f \colon \mathcal{X} \to \mathbb{R}$ is trained given a training set $\{(\boldsymbol{x}_{i}, y_{i})\}^{n}_{i=1} \subset \mathcal{D}$ with a target function $g \colon \mathcal{X} \to \mathbb{R}$.
  Also, a training distribution $\mathcal{D}$ is defined by the following formulation.
  \begin{equation*}
    \mathcal{D} = \{(\boldsymbol{x}, y) \mid \min{\{y_{i}\}^{n}_{i=1}} \leq y \leq \max{\{y_{i}\}^{n}_{i=1}} \}
  \end{equation*}
  Let $ \ell \colon \mathbb{R} \times \mathbb{R} \to \mathbb{R}$  be a loss function and $\mathcal{P}$ be a distribution over $\mathcal{X} \setminus \mathcal{D}$. Then extrapolation is a task that model $f$ learns the distribution $\mathcal{P}$ by minimizing extrapolation error $\mathbb{E}_{\boldsymbol{x} \sim \mathcal{P}}\left[ \ell \left( f(\boldsymbol{x}), g(\boldsymbol{x}) \right) \right]$
\end{dfn}


\subsection{Hypothesis \label{sec:hypothesis}}
In this research, two hypotheses regarding the effects of self-supervised pretraining on extrapolation will be verified. 
We give an overview of the hypotheses and the technical reasons why they are made.

\subsubsection{H1: Is self-supervised pretraining effective for extrapolation? \label{sec:h1}}
First of all, we hypothesized that self-supervised pretraining itself is effective for extrapolation.
There are following two reasons for hypothesizing in this way.
Firstly, there are several results based on previous research that self-supervised learning and pretraining make models robust \cite{Hendrycks2019-kl, Mohseni2020-ef, Hendrycks2020-ij}. 
It would be of great merit to be able to acquire such expressions with various types of data.
Secondly, it is possible to incorporate non-linearity by pretraining. 
As explained in section \ref{sec:extrapolation}, MLP with ReLU cannot predict non-linearity outside the training distribution.
As a solution to this, Xu et al.(2021) \cite{xu2021neural} proposed to use representation learning with a different task before fine-tuning.
Based on the paper, we assume that non-linearity can be given to the model by performing self-supervised pretraining. 



\subsubsection{H2: Does utilization of unlabeled data help extrapolation? \label{sec:h2}}
As a second hypothesis, we insist that extrapolation can be improved by utilizing unlabeled data in self-supervised pretraining.
It can be expected that significant performance degradation on unlabeled data occurs since training by only labeled data leads to overfitting to them.
This expectation arises from one assumption that the unlabeled data holds important information which labeled data don't have in our definition.

Therefore, we hypothesized that it might be possible to reduce overfitting by treating both labeled and unlabeled data as inputs for self-supervised pretraining.
It can handle unlabeled data since it requires not labels but graph structures.

\subsection{Experiment Design \label{sec:design}}

\begin{figure}[t]
  \centering
  \begin{minipage}[t]{\hsize}
    \centering
    \includegraphics[width=\hsize]{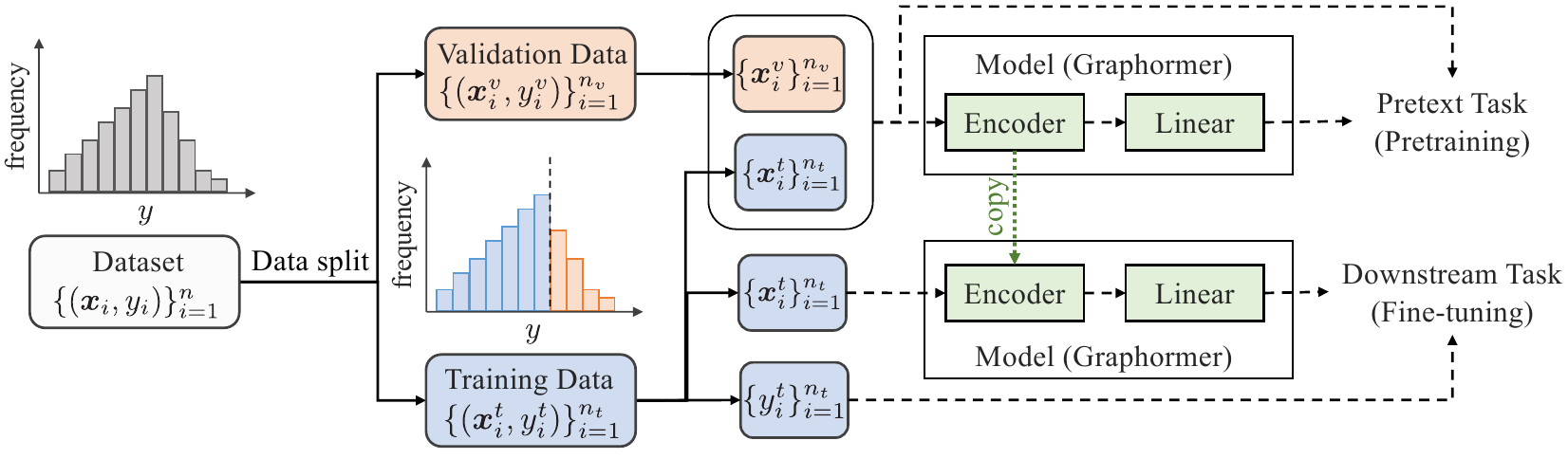}
    \caption{
      An overview of our experiment when models are trained by self-supervised pretraining and fine-tuning. The flow is illustrated from the data split to the downstream task.    }
    \label{fig:overview}
  \end{minipage}
\end{figure}

Firstly, we give an overview of the flow when using self-supervised pretraining in our experiment in \figref{overview}.
Training models in this experiment consists of two steps: self-supervised pretraining, and fine-tuning.
Also, there are three important elements:
1) the model for predicting property from the molecular graph;
2) the way to train the model;
3) the method of data split;
We will introduce all of them in this section.

\subsubsection{Model\label{sec:model}}

In this research, graphormer \cite{Ying2021-ye} based on a transformer is adopted which consists of encoders and a linear layer.
We chose the model since it is currently near the state of the art in molecular property prediction.

\subsubsection{Training \label{sec:training}}

we first train the model by self-supervised pretraining with training and validation data
This step aims to train the encoder by either of three specific pretext tasks.
After that, fine-tuning is conducted using labeled data by minimizing Mean Absolute Error (MAE).

Then, the three pretext tasks are
1) node-level masking task (attribute prediction);
2) geometric structure prediction (structure prediction);
3) motif prediction (attribute + structure prediction);
We chose them because they can cover various sizes of trends in molecular graphs.
As node attributes are one of the minimum units in a graph, the node-level masking task is expected to enable the encoder to learn micro tendencies.
The geometric structures prediction also can give connectivity information.
Motif prediction is a combination of attribute and structure prediction.
We provide detailed explanations for each of the three tasks.


\textbf{Task1 : Node-level masking task}.
This tasks is a classification problem, predicting node attributes of masked parts given the input with partially masked nodes.
The nodes are randomly selected before starting training, and the same masks are consistently used during the training.
As for the masking rate, one node in one molecular graph is masked.
The model is trained to minimize the cross-entropy loss function.

\begin{figure}
  \centering
  \begin{minipage}[t]{0.49\hsize}
    \centering
    \includegraphics[width=0.9\hsize]{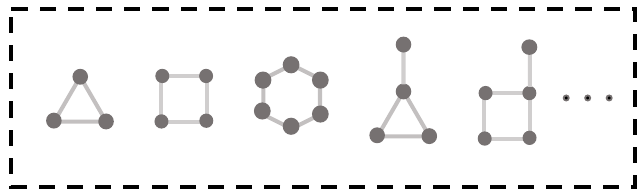}
    \caption{
      An example of a structure vocabulary and a structure label.
      The vocabulary consists of 141 structures.
    }
    \label{fig:structure_vocab}
  \end{minipage}
  \hfill
  \begin{minipage}[t]{0.48\hsize}
    \centering
    \includegraphics[width=0.9\hsize]{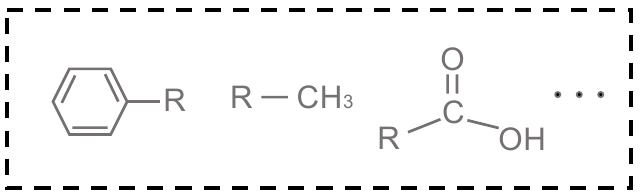}
    \caption{
      An example of a motif vocabulary and a motif label.
      The vocabulary consists of 167 motifs from MACCS keys \cite{Durant2002-rj}.
    }
    \label{fig:motif_vocab}
  \end{minipage}
\end{figure}

\textbf{Task2 : Geometric structure prediction}.
The task is to predict geometric structures only.
In other words, a model judges the presence of connections between nodes without considering attributes.
To achieve this, we enumerated 141 fully connected graphs whose nodes consist of 3 to 6 vertices.
Then, we selected 83 structures that at least one molecule includes and created the structure vocabulary as illustrated in \figref{structure_vocab}.
We encoded information of whether certain structures are included in a molecule or not as an 83-dimensional label.
The labels are made by using function \emph{subgraph\_is\_isomorphic} in NetworkX \cite{hagberg2008exploring}.
Also, since this problem can be treated as a multi-label binary classification problem, we train a model to minimize the binary cross-entropy loss function.

\textbf{Task3 : Motif prediction}.
The last task is to predict the presence of motifs in a graph.
Motifs are graph structures with important functional features based on domain knowledge.
They contain not only structures but also attributes, and this task is a graph-level prediction.
This research uses the molecular structures determined by chemical domain knowledge called MACCS keys \cite{Durant2002-rj}. 
These motifs contain meaningful 167 substructures, such as -COOH and -OH, as seen from \figref{motif_vocab}.
As conducted in task2, each molecule is checked if it contains these motifs, creating a label for the 167-dimensional motifs.
These operations are performed by the function \emph{GetMACCSKeysFingerprint} provided by RDKit \cite{rdkit}.
In addition, we used the binary cross entropy loss function as the loss function.

\subsubsection{Data Split \label{sec:split}}
Cross-validation is the mainstream of conventional data-splitting methods.
However, this method is not suitable for generating validation data for evaluating extrapolation.
Therefore, we use the forward holdout validation proposed by Xiong et al. (2020) \cite{Xiong2020-qy} for preparing validation data.
The method splits dataset into training data $\{(\boldsymbol{x}^{t}_{i}, y^{t}_{i})\}^{n_{t}}_{i=1}$  and validation data $\{(\boldsymbol{x}^{v}_{i}, y^{v}_{i})\}^{n_{v}}_{i=1}$ so that $\max\left(\{y^{t}_{i}\}^{n_{t}}_{i=1}\right) <  \min\left(\{y^{v}_{i}\}^{n_{v}}_{i=1}\right)$.

\subsection{Experiment Protocol \label{sec:protocol}}

To validate the two hypotheses, we train models by the following 7 training methods.

\textbf{No pretraining (baseline)}. 
As a baseline, models are trained by only fine-tuning with training data.
Then their interpolation and extrapolation performance is evaluated by holdout validation and forward holdout validation.

\textbf{Pretraining without validation data (task1, task2, task3)}. 
To check on whether utilizing unlabeled data in self-supervised pretraining is effective or not, the models are firstly pretrained with only training data.
As self-supervised pretraining, task1, task2, and task3 are used. 
After that, they are fine-tuned by training data.
Their extrapolation performance is evaluated by forward holdout validation.

\textbf{Pretraining with validation data (task1, task2, task3)}. 
As opposed to the above methods, in this training method, we use both training and validation data when pretraining. The rest procedures are the same as above.

\section{Evaluation \label{sec:evaluation}}
Firstly, we explain experimental settings throughout the three experiments.
We use PCQM4Mv2 for molecular property prediction provided by OGB-LSC \cite{Hu2020-vj} as a dataset.
It provides molecular graphs and HOMO LUMO gaps, and the model has to predict the gap values.
The model is graphormer, and we set the number of encoder layers to 12, the dimensions of feature vectors to 512, and adopted AdamW provided by PyTorch as the optimizer.
As for the experimental environment, all experiments were performed using 4 GPUs, and the model is trained with 30 epochs for pretraining and 80 epochs for fine-tuning.
Evaluation metrics are the averages of three runs in each method.

As an overview of the experiments, there are mainly three important findings as follows.
\begin{itemize}
  \item Graphormer trained by only fine-tuning can't extrapolate well. Especially it can't output over a certain property value, which would be a piece of evidence that it can't learn nonlinearity.
  \item Self-supervised pretraining improves performance to predict ranks of property values slightly, although models can't predict the property values exactly.
  \item The impact on utilization of unlabeled data can be seen from differences of rank correlations between with and without unlabeled data when pretraining.
\end{itemize}

\subsection{Analysis of MAE}
We evaluated all of the performances by Mean Absolute Error.
Firstly, we compare the result of interpolation and extrapolation in the baseline.
The average of minimum MAEs in interpolation reaches 0.091, whereas that in extrapolation does 0.692.
Additionally, the minimums in extrapolation are recorded from the 1st epoch to the 10th epoch.
It means that, in extrapolation, the model is considered to be poorly learned.

\begin{figure}
    \begin{minipage}[b]{0.5\hsize}
        \centering
        \begin{minipage}[]{0.48\hsize}
            \centering
            \includegraphics[width=\hsize]{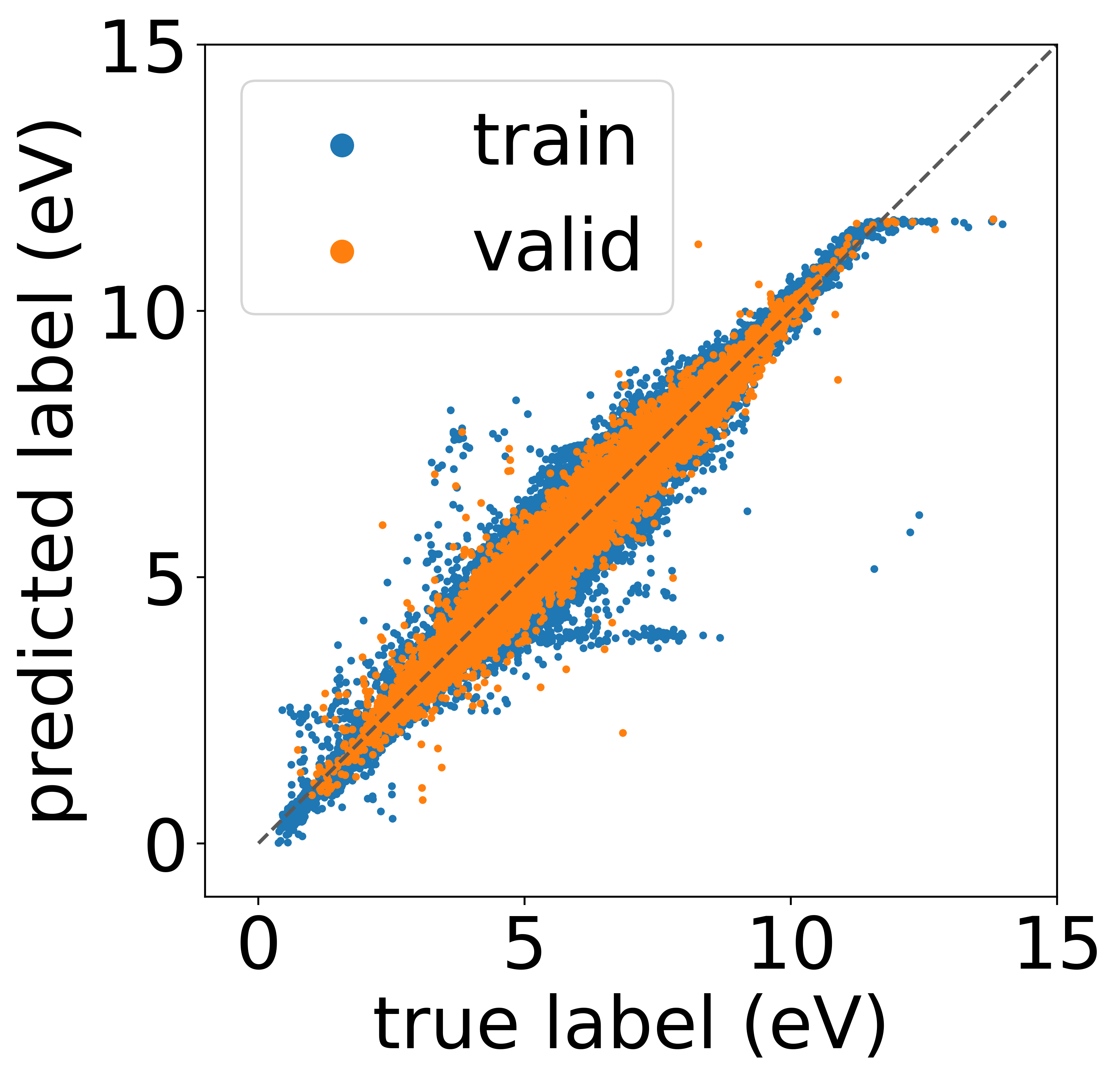}
            \subcaption{
                interpolation
            }
            \label{fig:1st_id}
        \end{minipage}
        \hspace{0.01\hsize}
        \begin{minipage}[]{0.48\hsize}
            \centering
            \includegraphics[width=\hsize]{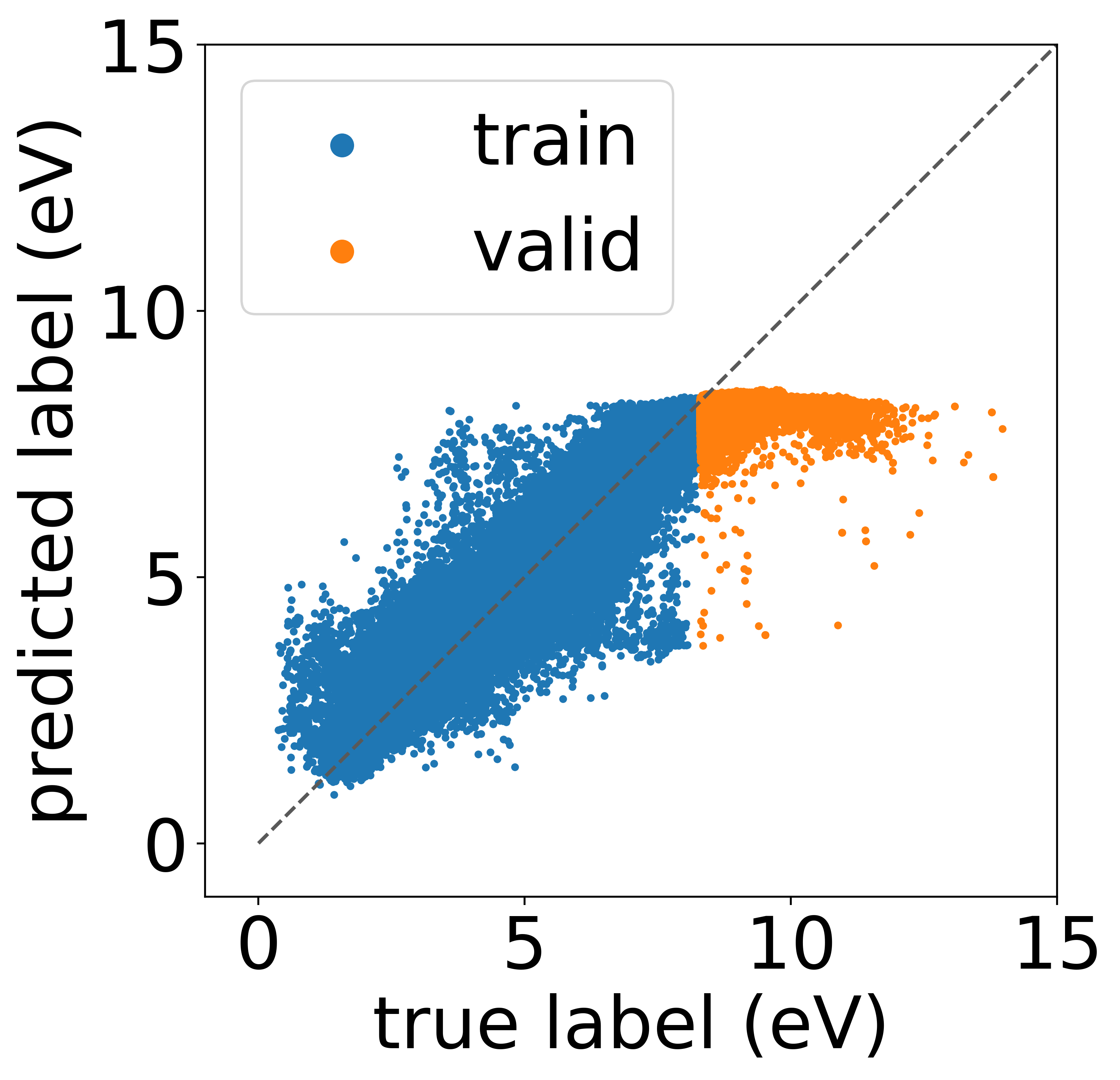}
            \subcaption{
                extrapolation
            }
            \label{fig:1st_ood}
        \end{minipage}
        \caption{
            Relationships between true and predicted label when training by baseline. The dotted black line is ideal.
        }
        \label{fig:1st_value}
    \end{minipage}
    \hfill
    \begin{minipage}[b]{0.48\hsize}
        \centering
        \includegraphics[width=0.99\hsize]{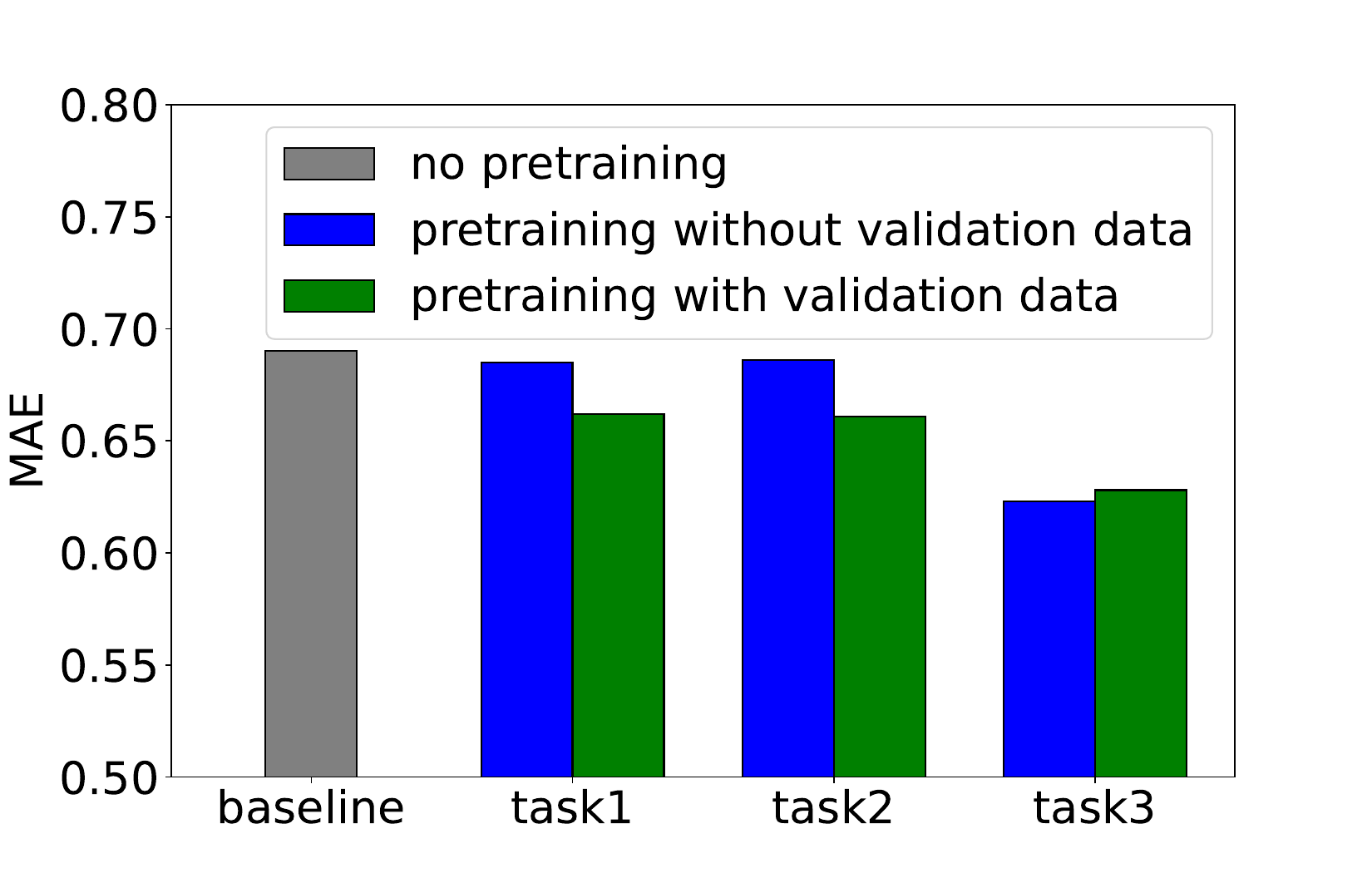}
        \caption{
          The result of Mean Absolute Error in extrapolation.
        }
        \label{fig:mae}
  \end{minipage}
\end{figure}

To investigate the issue in more detail, we explain \figref{1st_value} which shows the relationship between the predicted value and the label.
First of all, in interpolation, the validation data is plotted along the black dotted line.
On the other hand, in extrapolation, the validation data are far apart, and the predicted value hardly exceeds a certain value.
One of the reasons for this result may be that there are no data with labels greater than about 8 in the training data.
However, since a large number of predicted values exist around 8, the model seems to be able to predict the tendency that the validation data is relatively high in the dataset.

\figref{mae} shows the mean absolute error in extrapolation with 7 training methods.
Considering baseline and pretraining with validation data, we can find that the MAEs of task1, 2, and 3 are lower than that of the baseline method
When compared to the baseline, the values decreased by 4.91\% for method1, 3.47\% for method2, and 9.68\% for method3.
However, these MAEs are much larger than the interpolation MAEs, and the extrapolation performances are far behind the interpolation.

When focusing on the result of pretraining with and without unlabeled data, tasks 1 and 2 with unlabeled data are almost 3\% better than without them.
In contrast, the MAE of task 3 increased by 0.8\% when unlabeled data is used in pretraining.

In contrast, all of the MAEs are significantly worse than that in interpolation.
It means that self-supervised pretraining can't help models extrapolate exact values.
Therefore, in terms of predicting exactly, the hypothesis h1 and h2 are false.

\begin{figure*}[t]
  \centering
  \begin{minipage}[t]{0.245\hsize}
    \centering
    \includegraphics[width= 0.95\hsize]{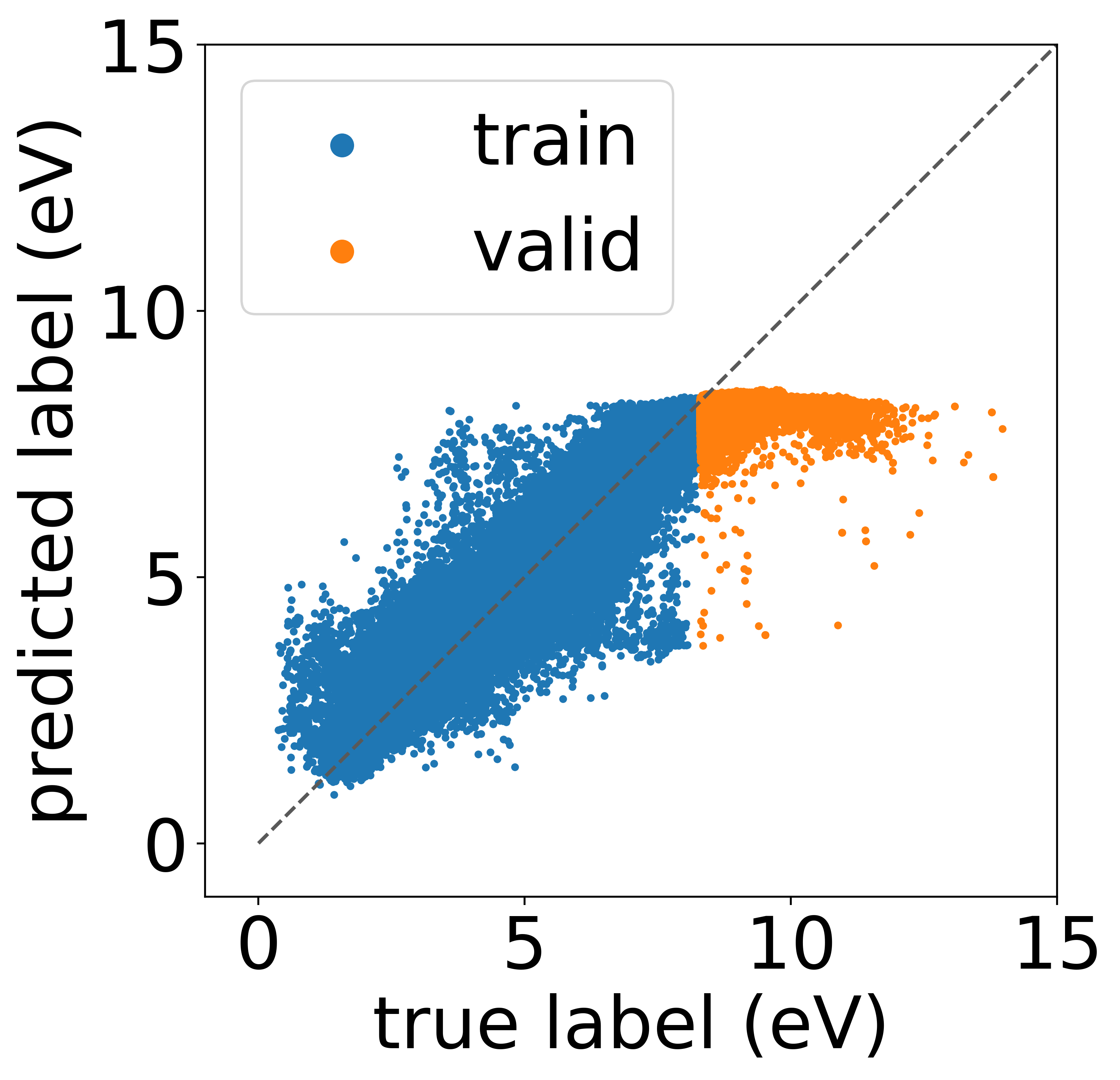}
    \subcaption{
      baseline
    }
    \label{fig:2nd_baseline}
  \end{minipage}
  \hfill
  \begin{minipage}[t]{0.245\hsize}
    \centering
    \includegraphics[width=0.95\hsize]{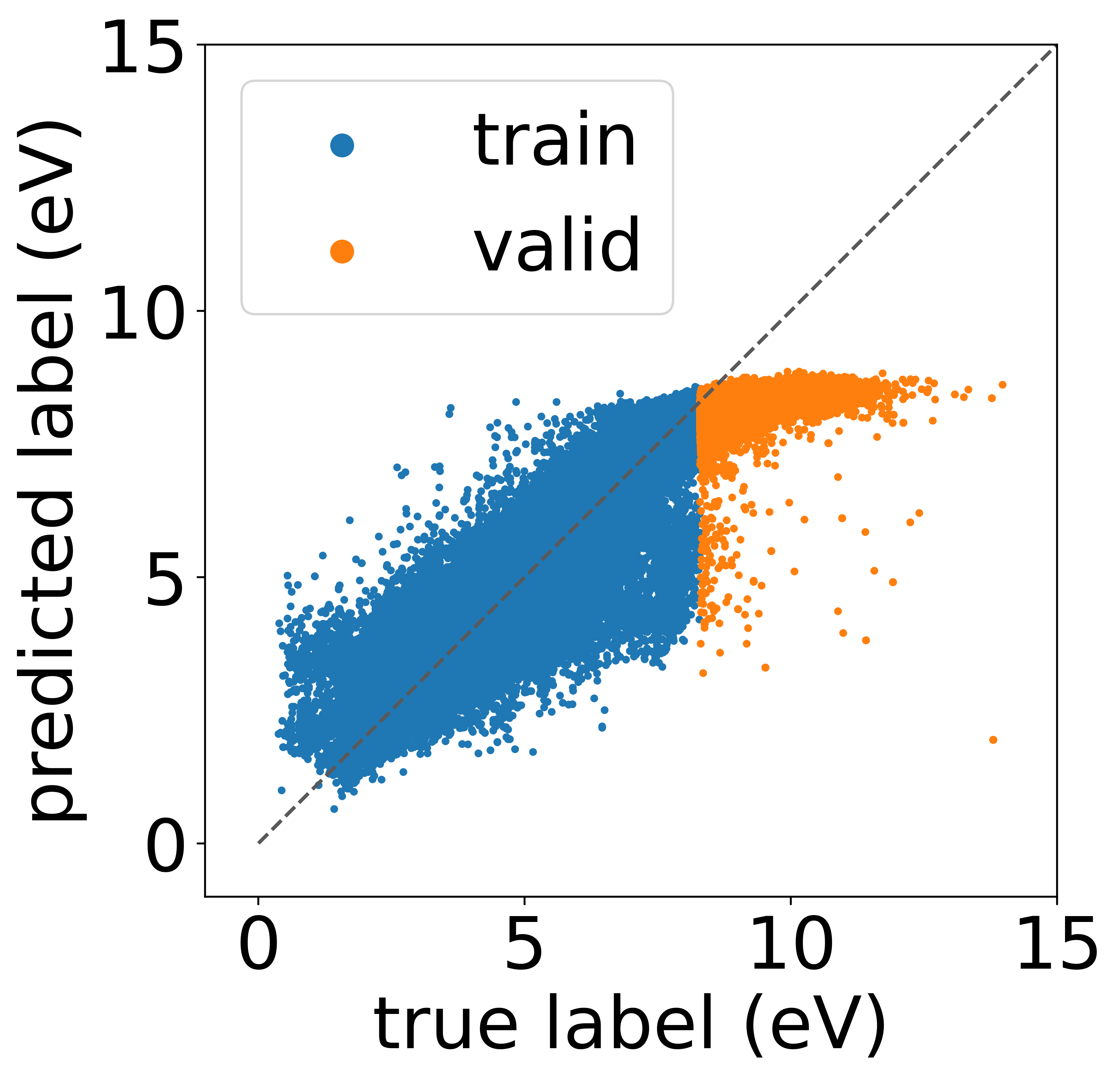}
    \subcaption{
      task1
    }
    \label{fig:2nd_method1}
  \end{minipage}
  \centering
  \begin{minipage}[t]{0.245\hsize}
    \centering
    \includegraphics[width= 0.95\hsize]{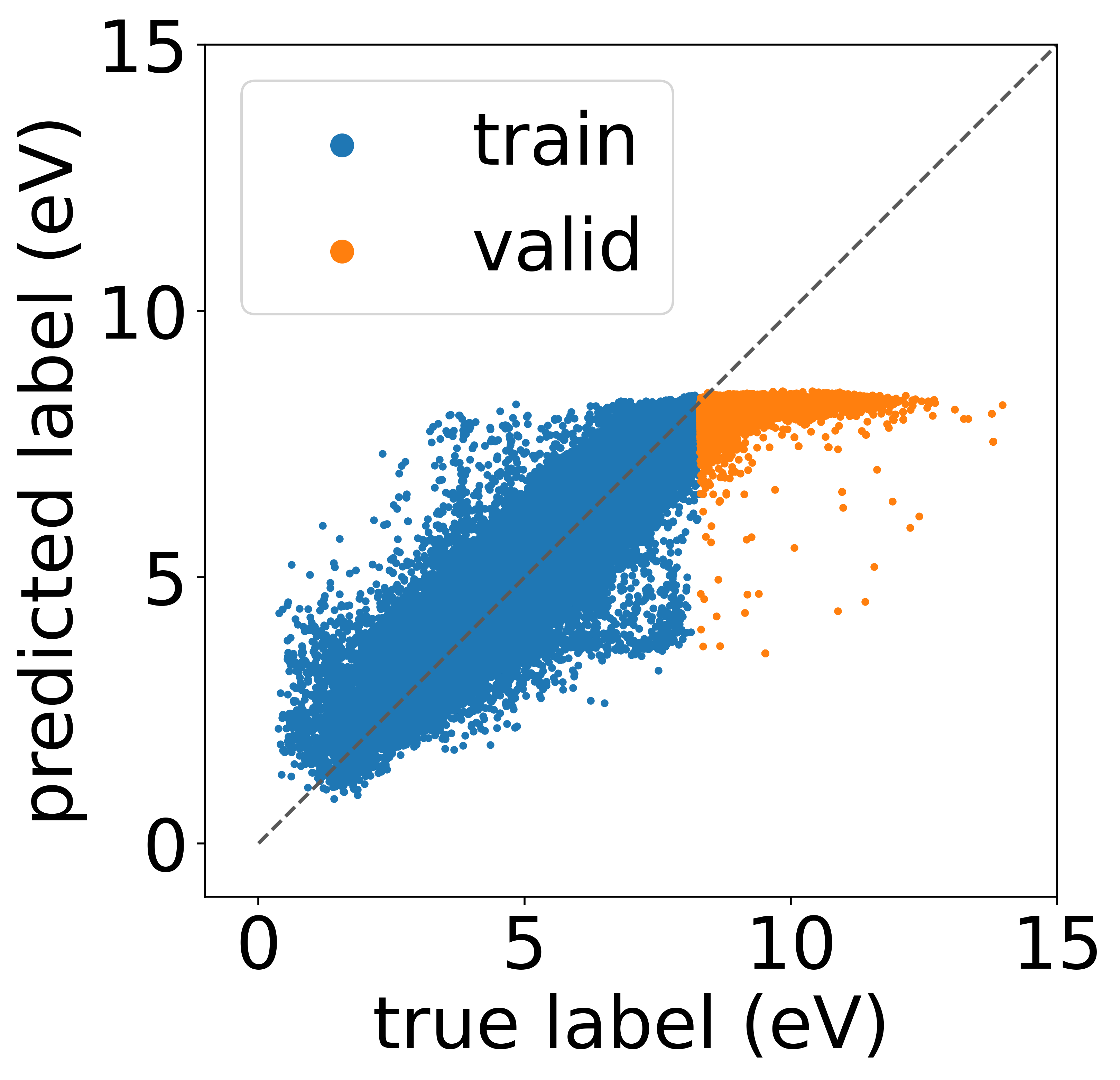}
    \subcaption{
      task2
    }
    \label{fig:2nd_method2}
  \end{minipage}
  \hfill
  \begin{minipage}[t]{0.245\hsize}
    \centering
    \includegraphics[width=0.95\hsize]{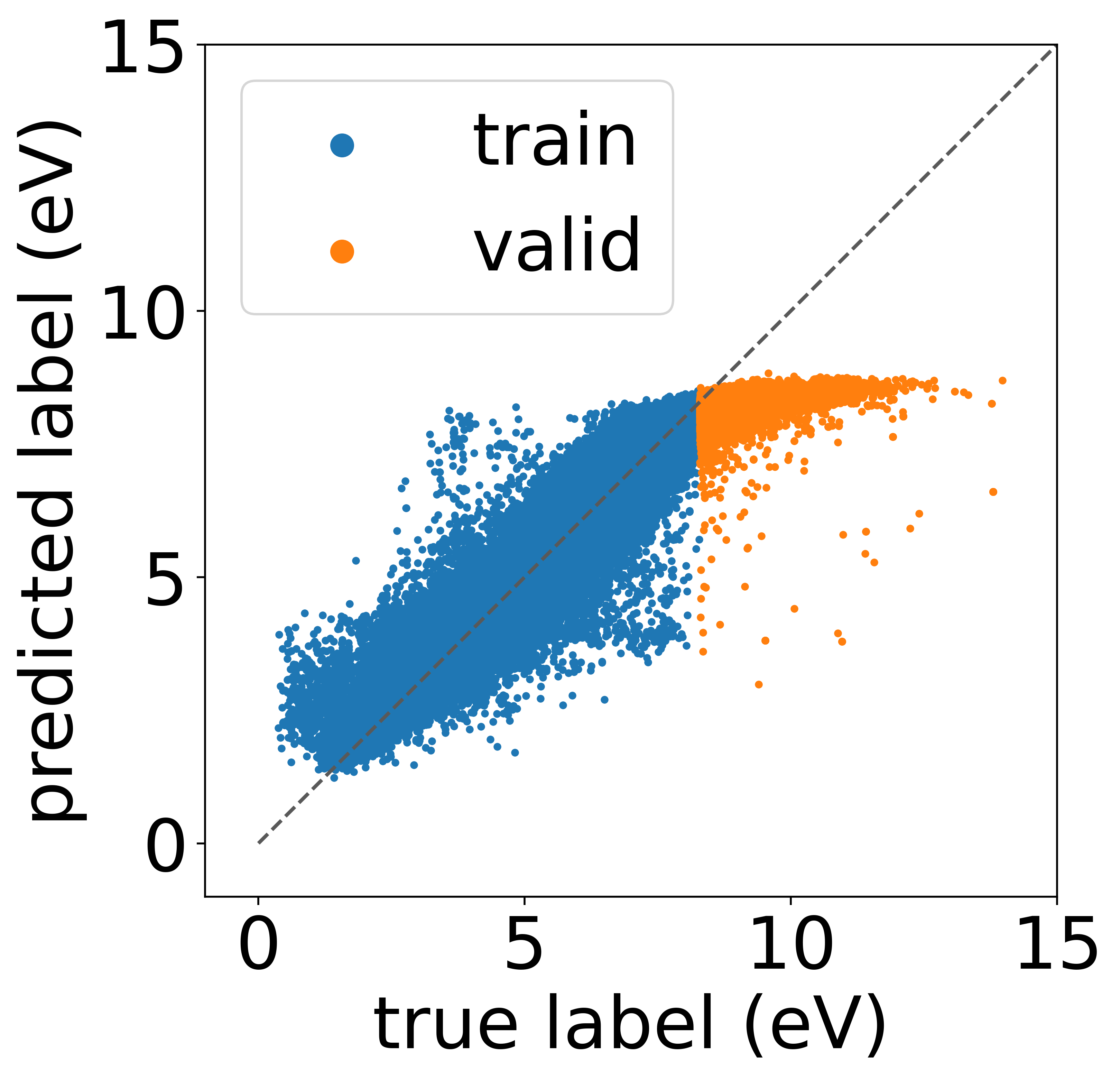}
    \subcaption{
      task3
    }
    \label{fig:2nd_method3}
  \end{minipage}
  \caption{
    The relationships between true label and predicted label in extrapolation. Task1, 2 and 3 is used in self-supervised learning with both training and validation data. The dotted black line is ideal.
  }
  \label{fig:2nd_value}
\end{figure*}

Also, we illustrate \figref{2nd_value} to show the relationship between labels and predicted values.
In these four figures, the models could not almost predict values higher than 8.
In contrast, the difference between them is that the validation data in the baseline figure is slightly tilted to the lower right, while in task1 and 2 is almost parallel, and in task3 slightly tilted to the upper right.
This fact suggests that pretraining may help the model to learn relative trends in the validation data.
From this point of view, we considered that MAE alone is insufficient as an evaluation metric for extrapolation in this study, and decide to conduct further analysis.

\subsection{Analysis of Rank}

To analyze the relative tendency of the predictions and find out the difference between pretraining and non-pretraining methods more clearly, we introduce the idea of ranking.

This is an index that indicates whether the ranking in descending order of physical property value is correctly predicted in the validation data or not.
In addition, in the field of materials chemistry, understanding this ranking also has a positive effect on improving the efficiency of material development, so we decided to adopt this index.
Furthermore, ranking is a discrete value, and it is expected that the plot will be more sparse than \figref{2nd_value}, making it easier to see the difference from the figure.

To evaluate it, the rankings of the labels and the predicted values are calculated in advance, and the correlation coefficient is used as the evaluation metrics.
The reason why we choose the metric is that since the number of validation data is about 70,000, the metric is suitable for grasping the macro trends related to rankings.

\begin{figure*}
  \centering
  \begin{minipage}[b]{0.48\hsize}
    \centering
    \includegraphics[width=\hsize]{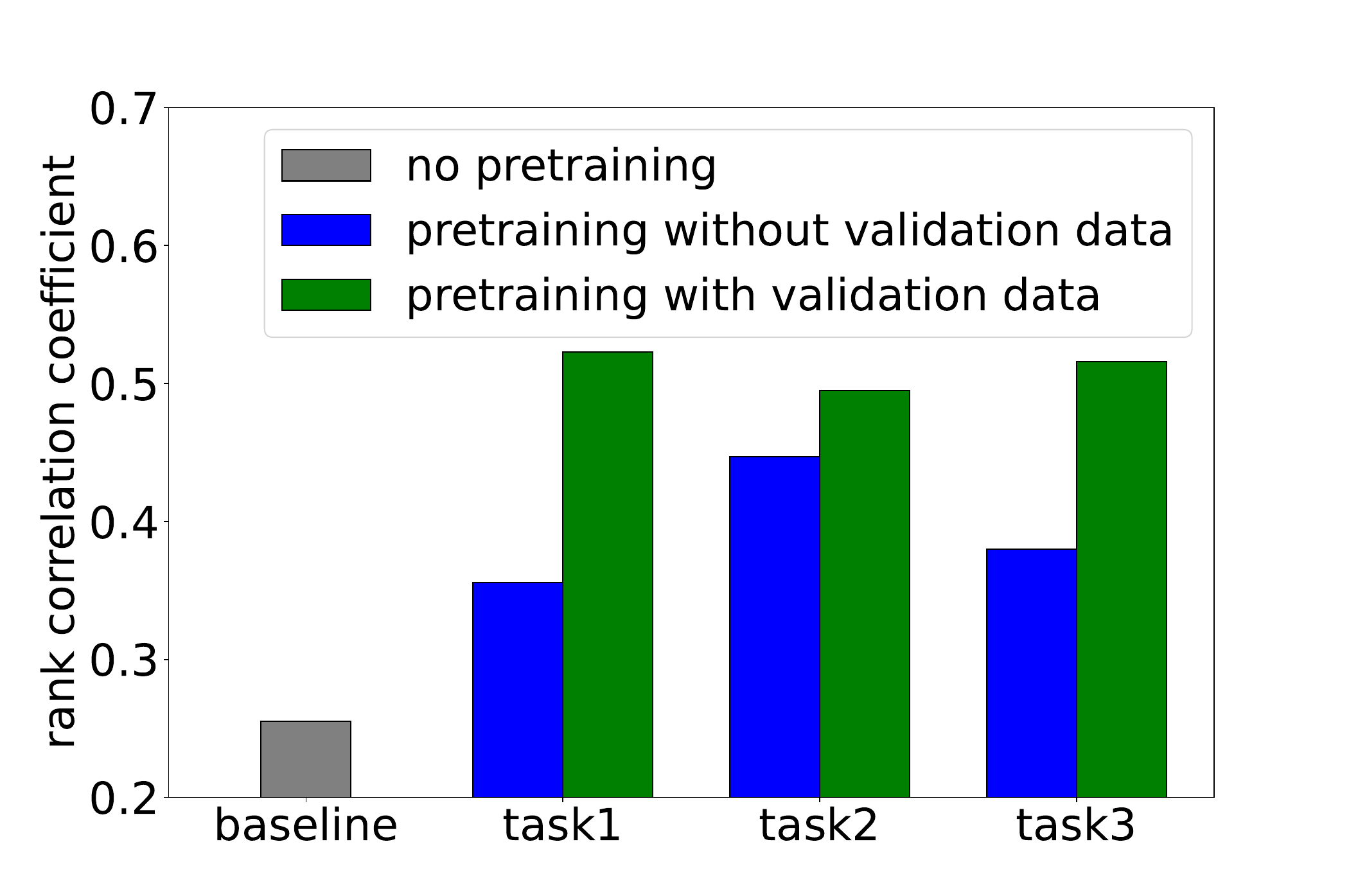}
    \caption{
      The result of rank correlation coefficients when the MAEs are the best in extrapolation.
    }
    \label{fig:rank}
  \end{minipage}
  \hspace{0.01\hsize}
  \centering
  \begin{minipage}[b]{0.48\hsize}
    \centering
    \includegraphics[width=\hsize]{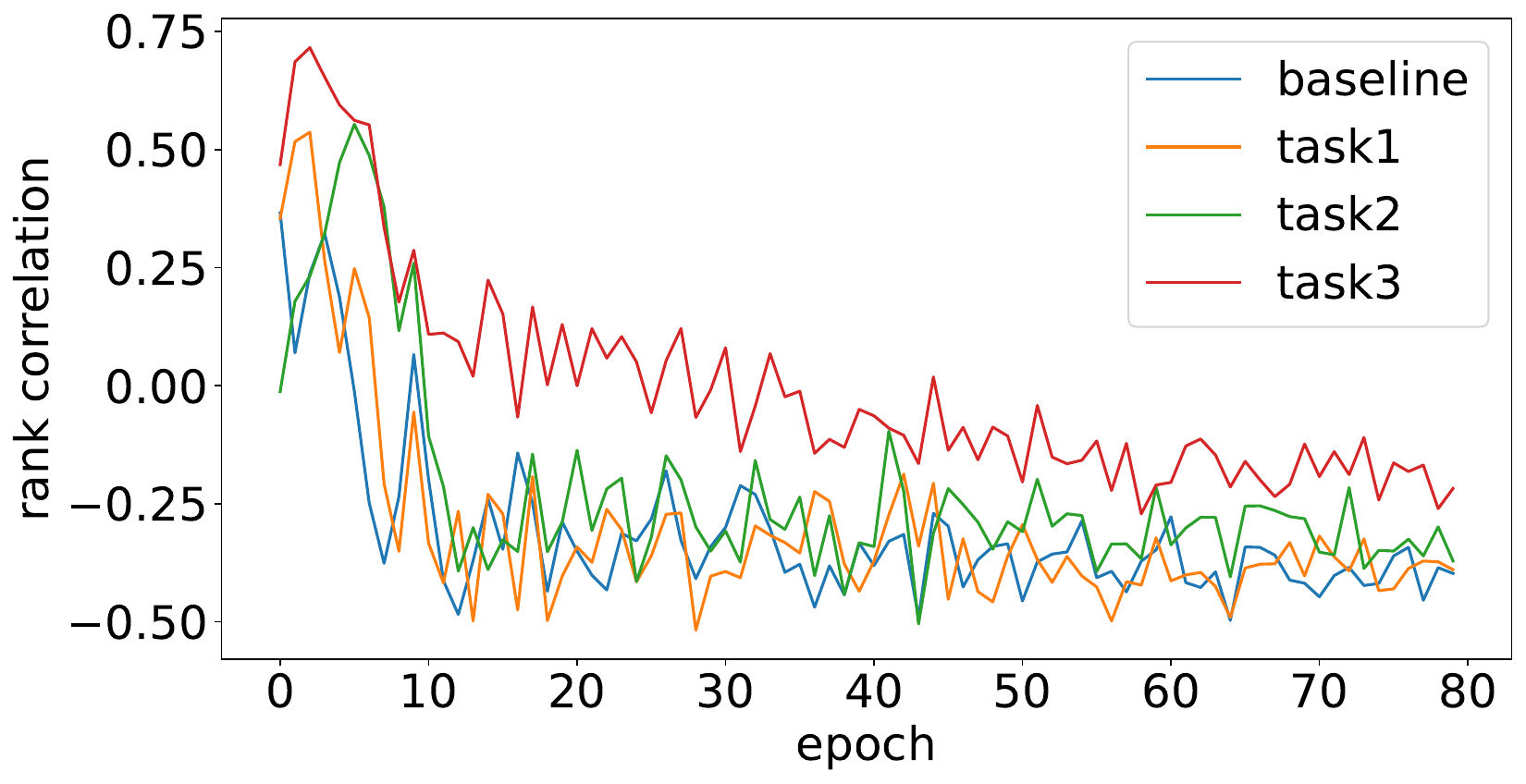}
    \caption{
        Behaviors of ranking correlation coefficients in extrapolation.
        Task1, 2, and 3 are used in self-supervised learning with both training and validation data.
    }
    \label{fig:2nd_rank_behavior}
  \end{minipage}
\end{figure*}

\figref{rank} illustrates the ranking correlation coefficients in extrapolation with 8 training methods.
What we can see from this figure is that the correlation coefficients for task1, 2, and 3 when using validation data are much closer to 1.0 than the baseline.
Specifically, the figure of task3 is 0.516, which is the best result.
On the other hand, the baseline is 0.255, which is the lowest value.
Correlation coefficients in the range from 0.4 to 0.7 are considered to have some degree of positive correlation, while if it is between -0.2 and 0.2 there is almost no correlation.
It can be said that self-supervised pretraining enhanced the ability to judge whether certain data is greater than other data in the validation data.

Then, for checking hypothesis h2, we contrast pretraining with and without validation data.
The result is that the metrics in all of the tasks decrease from almost 0.5 to 0.4 when removing validation data from self-supervised pretraining.
Therefore, it can be concluded that validation data has a good influence on extrapolation performance in terms of predicting the relative tendency.

Also, we calculated the correlation coefficients for each epoch and gain \figref{2nd_rank_behavior} showing the change through training.
From this figure, we can see the fact that by 10 epochs the correlation coefficient values fall into negative for most methods.
Especially in the first 5 epochs, the correlation coefficient is very high, and the method using pretraining has a value of 0.4 or higher, but it can be seen that it drops sharply after the epoch.
This phenomenon might be because, through training, a model that is too well-fitted to the training data fails to predict label values so high that the training data doesn't include it.
The issue is a significant limitation in our study and should be well analyzed.


\section{Conclusion}
In this paper, we have proposed the experimental framework for evaluating the impacts of self-supervised pretraining on extrapolation by dividing distributions based on labels and proved the hypotheses empirically using the benchmark of molecular property prediction.
Firstly, we have organized existing definitions of extrapolation and pointed out the necessity to decide distributions by a label for material development.
In the experiments, three methods with pretraining were demonstrated to be superior to baseline methods without pretraining by evaluating MAE, ranking.
Although the pretraining with both the labeled and unlabeled data couldn't improve the performance of extrapolating exact values, it could help to learn the tendency of unseen relative distributions.
The extrapolation ability could contribute to improving the efficiencies to search for new materials.

\section{Limitation}
Up to now, we explained the contributions of this work, but there are some limitations that should be solved in the future as follows.

\textbf{Impacts of pretraining}
As explained in \figref{2nd_rank_behavior}, extrapolation performances were improved in early epochs, and after that, the model seemed to be overfitted to training distribution.
Therefore, the impacts should last as long as possible.

\textbf{Imbalance of data split}
In our study, the dataset is split by forward holdout validation for extrapolation.
This method doesn't consider balance in distributions of training and validation data, there are still issues related to the way to divide the dataset for validation.

\textbf{Lack of downstream tasks}
As we have conducted only one benchmark OGB-LSC, there are possibilities that self-supervised pretraining is not effective in other benchmarks.
Therefore, future work has to verify the hypothesis in various downstream tasks.

\bibliographystyle{plain}
\bibliography{neurips_2023}

\newpage


\appendix
\section{Detail of Dataset}
In our study, we utilized the PCQM4Mv2 dataset, proposed as a benchmark for molecular property prediction in OGB-LSC \cite{Hu2020-vj}. This dataset incorporates the two-dimensional graph structure and three-dimensional coordinates of molecules as inputs; however, we exclusively focused on the former. 

\begin{figure}[h]
  \centering
  \begin{minipage}[h]{\hsize}
    \centering
    \includegraphics[width=0.7\hsize]{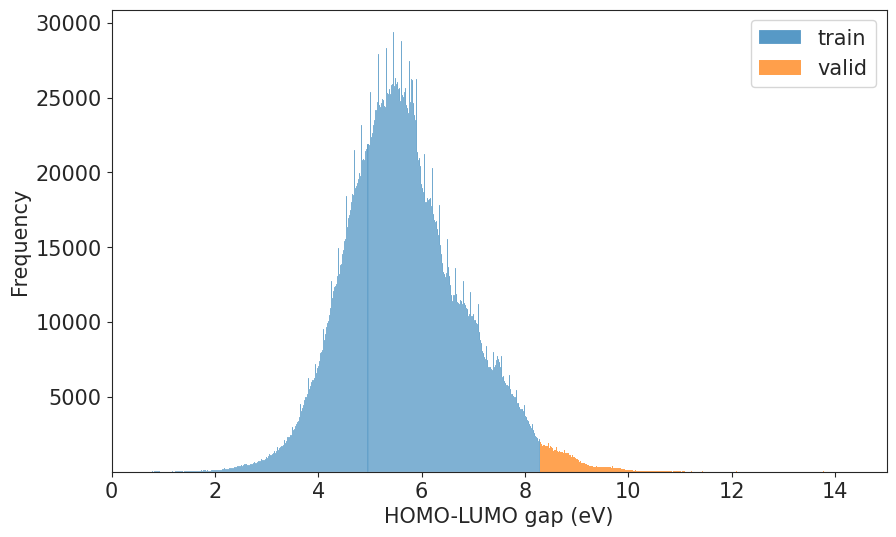}
    \caption{
        Distribution of HOMO-LUMO gap in train and validation data.
        Although there are data with an energy gap over 15eV, the figure is limited between 0eV and 15eV.
    }
    \label{fig:distribution}
  \end{minipage}
\end{figure}

This task involves predicting the HOMO-LUMO gap, and the distribution of its values is illustrated in \figref{distribution} In this study, we exclusively utilized the training and validation data for our analysis. The dataset comprises a total of 3,746,619 molecules, which are divided into train/validation/test-dev/test-challenge with a ratio of 90/2/4/4. For our study, we selected forward holdout validation to split the data, with the left side of the distribution representing the training data and the right side representing the validation data, maintaining a ratio of 90/2.

\section{Implementation Detail}
\subsection{Model}

This research uses graphormer as the model and re-implements it using PyTorch for our experiment. It firstly embeds node and edge features from the graph structure whose dimensions are set to 512 and 128 respectively. Additionally, the encoder layer, which includes an attention mechanism, is configured with 12 layers, and the dropout rate is set at 10\%.

\subsection{Training}

As described in the section of experiment design, we adopted a self-supervised pretraining approach as our training method. 
The details of this self-supervised pretraining are outlined in Section 3.3.2. 
Our training methodology consists of seven approaches, including one approach without pretraining, three approaches with pretraining not using validation data, and three approaches with pretraining using validation data.
The pretraining phase runs for 30 epochs, followed by fine-tuning for 80 epochs. An initial learning rate is 2$e$-4, controlled by the \textit{get\_polynomical\_decay\_schedule\_with\_warmup} function \footnote{https://github.com/huggingface/transformers/blob/main/src/transformers/optimization.py}, and the AdamW is used as optimizer.


\end{document}